\pdfoutput=1

\documentclass[11pt]{article}

\usepackage[]{acl}

\usepackage{times}
\usepackage{latexsym}

\usepackage[T1]{fontenc}

\usepackage[utf8]{inputenc}

\usepackage{microtype}

\usepackage{inconsolata}

\usepackage{arydshln}
\usepackage{bbold}
\usepackage{makecell}
\usepackage{booktabs} 
\usepackage{bbm}
\usepackage{amsmath}
\usepackage{graphicx,xcolor}
\usepackage{subcaption}
\usepackage{algorithm}
\usepackage{algpseudocode}

\title{Sampling-based Pseudo-Likelihood for Membership Inference Attacks}

\author{Masahiro Kaneko$^{1,2}$ \quad
        Youmi Ma$^{2}$\thanks{Masahiro Kaneko and Youmi Ma contributed equally to this work.} \quad
        Yuki Wata$^{3}$
        \quad
        Naoaki Okazaki$^{2}$ \\
        $^1$MBZUAI \quad
        $^2$Tokyo Institute of Technology \quad
        $^3$The University of Tokyo\\
        {\tt Masahiro.Kaneko@mbzuai.ac.ae} \quad
        {\tt youmi.ma@nlp.c.titech.ac.jp} \\
        {\tt uk1@is.s.u-tokyo.ac.jp} \quad
        {\tt okazaki@c.titech.ac.jp}
}

\begin{document}
\maketitle
\begin{abstract}
Large Language Models (LLMs) are trained on large-scale web data, which makes it difficult to grasp the contribution of each text. This poses the risk of leaking inappropriate data such as benchmarks, personal information, and copyrighted texts in the training data.
Membership Inference Attacks (MIA), which determine whether a given text is included in the model's training data, have been attracting attention.
Previous studies of MIAs revealed that likelihood-based classification is effective for detecting leaks in LLMs.
However, the existing methods cannot be applied to some proprietary models like ChatGPT or Claude 3 because the likelihood is unavailable to the user.
In this study, we propose a Sampling-based Pseudo-Likelihood (\textbf{SPL}) method for MIA (\textbf{SaMIA}) that calculates SPL using only the text generated by an LLM to detect leaks.
The SaMIA treats the target text as the reference text and multiple outputs from the LLM as text samples, calculates the degree of $n$-gram match as SPL, and determines the membership of the text in the training data.
Even without likelihoods,  SaMIA performed on par with existing likelihood-based methods.\footnote{Our code is available at: \url{https://github.com/nlp-titech/samia}}
\end{abstract}

\section{Introduction}

Large Language Models (LLMs)  bring about a game-changing transformation in various services used on a daily basis~\cite{NEURIPS2020_1457c0d6,touvron2023llama}.
The pre-training of LLMs relies on massive-scale web data of mixed quality~\cite{Zhao2023ASO}.
While pre-processing such as filtering is applied to construct as clean datasets as possible, it is unrealistic to remove everything undesired~\cite{Almazrouei2023TheFS}.
There is a risk of unintentionally leaking benchmark data, copyrighted texts, or personal information into the pre-training data~\cite{Kaneko2024ALL}.
The leakage of benchmark data can lead to an overestimation of LLMs' capabilities, preventing an appropriate assessment of their true performance~\cite{Yu2023BagOT,Zhou2023DontMY}.
Additionally, LLM generation based on copyrighted texts or personal information can result in serious violations of the law~\cite{Yeom2017PrivacyRI,Eldan2023WhosHP}.

Membership Inference Attacks (MIA) consider the task of determining whether a given target text is included in the training data of a  model~\cite{Shokri2016MembershipIA}.
Generally, because models are trained to fit the data,  a text included in the training data tends to exhibit a higher likelihood compared to ones unseen in the training data~\cite{Yeom2017PrivacyRI}.
Existing MIA studies rely on this idea and thus require the likelihood of a text computed by the model~\cite{DBLP:conf/uss/CarliniTWJHLRBS21,Inproc_lira,mattern-etal-2023-membership,shi2023detecting}.
It is impossible to apply the existing studies to the models that do not provide a likelihood.

\begin{figure}[t!]
    \centering
    \includegraphics[width=1\linewidth]{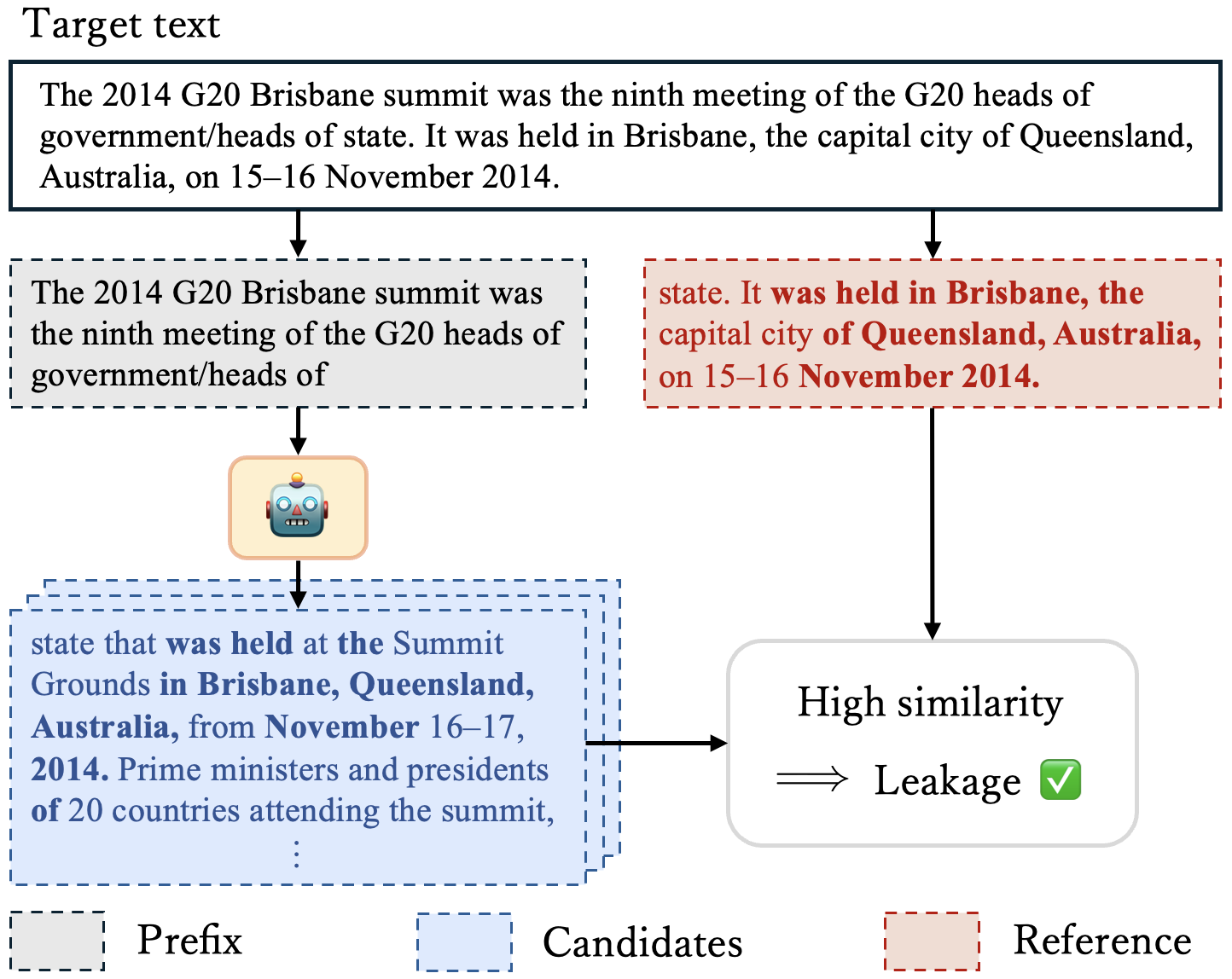}
    \caption{MIA using SPL based on the degree of $n$-gram between sampled candidate texts and a reference text.}
    \label{fig:SaMIA}
\end{figure}

When we are accessible to the training data of a model, it is straightforward to check the membership by directly searching for a given text in the training data
~\cite{zhang2022opt,biderman2023pythia,gpt-j}.
Therefore, the primary targets for applying MIA are LLMs without the training data disclosed~\cite{zhang2022opt,biderman2023pythia,touvron2023llama}.
Such LLMs mostly provide no 
access to likelihoods, which severely limits the practical applicability of likelihood-based MIA.
Moreover, as the commercial value of LLMs is increased, developers of recent powerful models like ChatGPT\footnote{\url{https://chat.openai.com/}}, Gemini\footnote{\url{https://gemini.google.com/app}}, and Claude 3\footnote{\url{https://claude.ai/chats}} are more reluctant to disclose the details of their training data.

In this paper, we propose a Sampling-based Pseudo Likelihood (\textbf{SPL}) method for MIA (\textbf{SaMIA}) that calculates SPL using the match ratio of $n$-grams between the output texts sampled from an LLM and the target text.
\autoref{fig:SaMIA} shows the detection of a leaked text using SPL, based on the $n$-gram similarity between text samples and the reference text.
Providing the initial part of the target text to the LLM, SaMIA generates multiple continuations of the text by sampling.
Treating the generated sequences as candidate texts and the remaining part of the target text (after the given part) as the reference text, we calculate the degree of $n$-gram overlaps between the candidate texts and the reference text.
If the degree of $n$-gram overlaps exceeds a specified threshold, we regard that the LLM has been trained on the text.

We conducted experiments on MIA with four LLMs (GPT-J-6B, OPT-6.7B, Pythia-6.9B, LLaMA-2-7B) whose training data is publicly available.
The experimental results on Wikipedia articles showed that SaMIA can achieve performance on par with existing methods based on likelihood or loss values.
In addition, we introduce a leakage detection method that combines information content and SPL.
The method achieved the highest average score of all existing methods.
We also report the analysis results of the impacts of the number $n$ of $n$-gram, the number of text samples, and the length of the target text on the performance of SaMIA.
The performance of SaMIA is highest when using unigrams. We observed the improved performance by increasing the number of text samples and length of the target text.

\section{SaMIA}
\label{sec:samia}

\subsection{Definition of the MIA Task}
MIA is a binary classification task to determine whether a target text $x$ is included in the training dataset $\mathcal{D}_{\text{train}}$ of a model $f_\theta$.
The attacker's goal is to design an appropriate attack function $A_{f_\theta}: \mathcal{X} \to \{0, 1\}$ and to determine the truth value of $x \in \mathcal{D}_{\text{train}}$ for an instance $x$ in the text space $\mathcal{X}$.

\begin{algorithm*}[t]
\caption{\,Sampling-based Membership Inference Attacks}
\label{algo:SaMIA}
\small
    \begin{algorithmic}[1]
    \State \textbf{Input:} target text $x=(w_1, w_2, \ldots, w_T)$, language model $f_\theta$, number of samples $m$, length of $n$-gram $N$, threshold $\tau$, zlib compression flag $z$
    \State \textbf{Output:} is text $x$ included in the training data of $f_\theta$? (1 or 0)
    \State $x_\text{prefix}\leftarrow(w_1, w_2, \ldots, w_{\lfloor T/2 \rfloor})$
    \State $x_\text{ref}\leftarrow(w_{\lfloor T/2 \rfloor+1}, w_{\lfloor T/2 \rfloor+2}, \ldots, w_T)$
    \For{$j = 1$ \textbf{to} $m$}
    \State $x_\text{cand}^{\,j} \leftarrow f_\theta(x_\text{prefix})$ \Comment{The $j$-th candidate text $x_\text{cand}^{\,j}$ generated from $f_\theta$ using $x_\text{prefix}$ as the prompt}
    \EndFor
    \If{$z$ = 1}
        \State $\overline{\text{R}}_m \leftarrow \frac{1}{m}\sum_{j=1}^m\text{ROUGE-N}(x_{\text{cand}}^{\,j}, x_{\text{ref}}) \cdot \text{zlib}(x_{\text{cand}}^j)$ \Comment{Combine with zlib compression}
    \Else
        \State $\overline{\text{R}}_m \leftarrow \frac{1}{m}\sum_{j=1}^m\text{ROUGE-N}(x_{\text{cand}}^{\,j}, x_{\text{ref}})$
    \EndIf
    \If{$\overline{\text{R}}_m > \tau$}
        \State \Return 1
    \Else
        \State \Return 0
    \EndIf
    \end{algorithmic}
\end{algorithm*}

\subsection{SPL}
\label{sec:spl}

Our method is applicable under severer conditions than the previous studies, which requires the loss $\mathcal{L}$ or token likelihood $P_\theta$ of the model $f_\theta$.
In other words, our method can be applied to any LLMs because it uses only generated texts without the loss or likelihood.
The proposed method is formalized as follows. For a text $x = (w_1, w_2, \ldots, w_T)$ of length $T$ to be detected, we divide it into a prefix $x_\text{prefix} = (w_1, w_2, \ldots, w_{\lfloor T/2 \rfloor})$ and a reference text $x_\text{ref} = (w_{\lfloor T/2 \rfloor+1}, w_{\lfloor T/2 \rfloor+2}, \ldots, w_T)$ based on the number of words $T$.
The LLM then generates $m$ text samples (\textit{candidates} hereafter) $x_\text{cand}^j (j=1,\ldots,m)$ that continue from the prefix $x_\text{prefix}$. We use these candidates for MIA.

SaMIA judges that a text $x$ is included in the training data of the LLM if the candidate text $x_\text{cand}^j$ generated by the LLM has a high surface-level similarity to the reference text $x_\text{ref}$.
We use ROUGE-N~\cite{lin-2004-rouge} as the similarity metric, which measures the recall of $n$-grams in the reference text.
Given a candidate text $x_\text{cand}$ generated by the LLM and a reference text $x_\text{ref}$, we calculate the ROUGE-N score (ranging from 0 to 1) with \autoref{eq:rougen}:

{\small
\begin{equation}
\text{ROUGE-N}(x_\text{cand}, x_\text{ref}) = \frac{\sum_{\text{gram}_n\in x_\text{ref}} \text{Count}_\text{match}(\text{gram}_n)}{\sum_{\text{gram}_n\in x_\text{ref}}\text{Count}(\text{gram}_n)}
\label{eq:rougen}
\end{equation}}
Here, $n$ is the length of the $n$-grams.
The denominator is the total number of $n$-grams in the reference text, and the numerator is the total number of $n$-grams that overlap between the candidate text and the reference text.
For example, a ROUGE-1 score is high when the words generated by the LLM appear in the reference text.

We expect that SPL can appropriately capture the distribution of LLM's generations by empirically sampling texts from the LLM.
Let $W$ be a random variable defined by the LLM, and let $P(W=x)$ be the probability that $W$ takes the text $x$. Let $X^{(j)}_x$ be a random variable such that $X^{(j)}_x = 1$ if the $j$-th sampled sequence from the language model is $x$, and $X^{(j)}_x = 0$ otherwise.
\begin{equation}
X^{(j)}_x = \begin{cases} 
1 & \text{if the } j\text{-th sampled sequence is } x \\
0 & \text{otherwise} 
\end{cases}
\end{equation}
The expected value of $X^{(j)}_x$ is calculated as follows:
\begin{align}
\mathbb{E}[X_x^{(j)}] & = 1 \cdot P(X_x^{(j)} = 1) + 0 \cdot P(X_x^{(j)} = 0) \nonumber \\
& = P(X_x^{(j)} = 1)
\end{align}
Here, the probability that $X_x^{(j)} = 1$ is equal to the probability that the language model generates the text $x$, that is $P(W=x)$.
When the sample size $N$ is sufficiently large, the average of $X_x^{(j)}$, which is the relative frequency of the text $x$ in the $j$-th sample, will converge to the expected value by the law of large numbers, and become equivalent to $P(W = x)$.
\begin{align}
\lim_{N \to \infty} \frac{1}{N} \sum_{i=1}^{N} X^{(i)}_x & = \mathbb{E}[X^{(i)}_x] \nonumber \\ & = P(W = x)
\end{align}
Therefore, by sampling the texts generated by the LLM, the appearance frequency of sampled texts approximates the probability that the language model generates a text. 
Thus, calculating the similarity between sampled texts and a target text can reflect the output tendencies of the LLM.

\subsection{Detection by SaMIA Based on SPL}

SaMIA calculates the average ROUGE-N score between each candidate text $x_\text{cand}^j$ generated by the LLM and the reference text $x_\text{ref}$.
If this average exceeds a threshold $\tau$, the text $x$ is considered as a member of the training data of the LLM (\autoref{eq:SaMIA}).

\begin{equation}
A_{f_\theta}(x) = \mathbb{1}\left[\frac{1}{m}\sum_{j=1}^m\text{ROUGE-N}(x_\text{cand}^j, x_\text{ref})>\tau\right]\label{eq:SaMIA}
\end{equation}
Here, $\mathbbm{1}[\cdot]$ stands for an indicator function and $m$ is the size of the test dataset.
We set the threshold $\tau$ based on the development data.
The interpretation of this detection metric is straightforward: if the text $x$ were used in training, the LLM would generate many of the $n$-grams present in the reference text $x_\text{ref}$.

\subsection{Improving SaMIA Using the Information Content}

Existing methods such as PPL/zlib use the information content computed by zlib compression to evaluate the redundancy characteristics of the generated text~\cite{DBLP:conf/uss/CarliniTWJHLRBS21}.
Samples from unseen data in training tend to contain repetitive generation (e.g., \textit{``I love you. I love you...''}), and the information content of such samples after zlib compression is expected to be lower.
PPL/zlib uses the ratio of the perplexity of the sample $x$ and the bit length of $x$ after zlib compression, $\mathrm{zlib}(x)$, as the detection metric (\autoref{eq:ppl/zlib}).
\begin{equation}
A_{f_\theta}(x) = \mathbb{1}\left[\frac{\prod_{i=1}^n P_\theta(x_i|x_{1:i-1})^{-\frac{1}{n}}}{\mathrm{zlib}(x)} < \tau\right] \label{eq:ppl/zlib}
\end{equation}
Here, $P_\theta(x_i|x_{1:i-1})$ is the model's likelihood of generating token $x_i$ given $x_{1:i-1}$ and $\mathrm{zlib}(x)$ represents the entropy of the text $x$ in bits after zlib compression.

Since $\mathrm{zlib}(x)$ depends only on the character information of the text $x$, it can also be applied to SaMIA.
Additionally, SaMIA (\autoref{eq:SaMIA}) does not consider the presence of repetitive generation in the candidate texts $x_\text{cand}^j$. Therefore, combining it with zlib can be expected to improve performance (\autoref{eq:SaMIA*zlib}).

{\small
\begin{align}
& A_{f_\theta}(x) = \notag \\
& \mathbb{1}\Bigg[\frac{1}{m}\sum_{j=1}^m\text{ROUGE-N}(x_\text{cand}^j, x_\text{ref}) \cdot \text{zlib}(x_\text{cand}^j)>\tau\Bigg] \label{eq:SaMIA*zlib}
\end{align}
}
Algorithm~\ref{algo:SaMIA} presents the detail of the whole process of the proposed method.


\section{Experiments}

\subsection{Existing Methods}

We compare SaMIA with several existing methods to evaluate the effectiveness of our proposed method.
The competitors range from basic to edge-cutting MIA approaches as follows.

\paragraph{LOSS~\cite{DBLP:conf/csfw/YeomGFJ18}}
The most straightforward MIA method, attacking by thresholding the Negative Log-Likelihood (NLL) loss.
Given a target text $x$, if model $f_\theta$ yields an NLL loss lower than a threshold $\tau$, then $x$ is considered present in the training data of model $f_\theta$:
\begin{equation}
A_{f_\theta}(x)=\mathbbm{1}[\mathcal{L}(f_\theta, x)<\tau],
\label{eq:loss}
\end{equation}
where $\mathcal{L}(f_\theta, x)$ is the NLL loss of model $f_\theta$ on $x$.

\paragraph{Lowercase~\cite{DBLP:conf/uss/CarliniTWJHLRBS21}} The method extends \textbf{LOSS} by thresholding the difference between a target text $x$ and its lowercase version $x_{\text{lower}}$:
\begin{equation}
A_{f_\theta}(x)=\mathbbm{1}[\mathcal{L}(f_\theta, x)-\mathcal{L}(f_\theta, x_{\text{lower}})<\tau].
\label{eq:lowercase}
\end{equation}

\paragraph{PPL/zlib~\cite{DBLP:conf/uss/CarliniTWJHLRBS21}}
The method alleviates the effect of repeated substrings via zlib compression\footnote{\url{https://www.zlib.net/}}.
PPL/zlib decides if a target text $x$ is included in the training data of model $f_\theta$ by thresholding the ratio of the log of the perplexity and the zlib entropy (\autoref{eq:ppl/zlib}).

\paragraph{Min-$k$\% Prob~\cite{shi2023detecting}}
The method utilizes only a subset of each target text $x$ during an attack.
Specifically, tokens in $x$ are sorted in the ascending order of the log-likelihood, and the average log-likelihood of the first $k\%$ tokens, composing a token set $\mathcal{E}$, are used for detection:

{\small
\begin{equation}
A_{f_\theta}(x) = \mathbbm{1}\left[\frac{1}{|\mathcal{E}|}\sum_{x_i\in{\mathcal{E}}} \log{P_\theta(x_i|x_{1:i-1})}>\tau\right].
\label{eq:mink}
\end{equation}
}
Following findings in~\citet{shi2023detecting}, we set $k=20$ for Min-$k$\% Prob in our experiments.

\subsection{Settings} 

\begin{table*}[t]
    \centering
    \footnotesize
    \begin{tabular}{lp{0.6cm}cp{0.6cm}cp{0.6cm}cp{0.6cm}c}
    \Xhline{3\arrayrulewidth}
    & \multicolumn{2}{c}{GPT-J-6B} & \multicolumn{2}{c}{OPT-6.7B} & \multicolumn{2}{c}{Pythia-6.9B} & \multicolumn{2}{c}{LLaMA-2-7B} \\
    \cmidrule(lr){2-3} \cmidrule(lr){4-5} \cmidrule(lr){6-7} \cmidrule(lr){8-9}
    & AUC & TPR@10\%FPR & AUC & TPR@10\%FPR & AUC & TPR@10\%FPR & AUC & TPR@10\%FPR \\
    \midrule
    \multicolumn{5}{l}{\textit{Existing Methods (likelihood-dependent)}} \\
    LOSS     & 0.66 & 18.0 & 0.61 & 16.8 & 0.65 & 20.4 & 0.55 & 13.8 \\
    PPL/zlib & 0.66 & 21.3 & 0.62 & 17.5 & 0.66 & 19.6 & 0.56 & 16.1 \\
    Lowercase & 0.59 & 16.7 & 0.58 & 16.0 & 0.57 & 17.0 & 0.52 & 15.5 \\
    Min-K\% Prob & \textbf{0.69} & \textbf{26.8} & 0.64 & 25.8 & \textbf{0.68} & \textbf{26.6} & 0.54 & 10.7  \\
    \midrule
    \multicolumn{5}{l}{\textit{Proposed Methods (likelihood-independent)}} \\
    SaMIA & 0.64 & 19.9 & 0.66 & 31.8 & 0.64  & 20.4 & 0.60 & 16.4 \\
    SaMIA*zlib & 0.66 & 26.0  & \textbf{0.71} & \textbf{37.3} & 0.66  & 23.7 & \textbf{0.62} & \textbf{20.0}\\
    \Xhline{3\arrayrulewidth}
    \end{tabular}
    \caption{Macro-averages of AUC and TPR@10\%FPR of the proposed and existing methods, when performing MIA on various LLMs using WikiMIA. The best score of each column is \textbf{bolded}.}
    \label{tab:macro}
\end{table*}

\begin{table*}[t]
\centering
\footnotesize
\begin{tabular}{lp{0.4cm}p{0.4cm}p{0.4cm}p{0.4cm}p{0.4cm}p{0.4cm}p{0.4cm}p{0.4cm}p{0.4cm}p{0.4cm}p{0.4cm}p{0.4cm}p{0.4cm}p{0.4cm}p{0.4cm}p{0.4cm}p{0.4cm}}
\Xhline{3\arrayrulewidth}
&\multicolumn{4}{c}{GPT-J-6B} &\multicolumn{4}{c}{OPT-6.7B} & \multicolumn{4}{c}{Pythia-6.9B} & \multicolumn{4}{c}{LLaMA-2-7B} \\
\cmidrule(lr){2-5} \cmidrule(lr){6-9} \cmidrule(lr){10-13} 
\cmidrule(lr){14-17} 
Length  & \multicolumn{1}{c}{32} & \multicolumn{1}{c}{64} & \multicolumn{1}{c}{128} & \multicolumn{1}{c}{256} & \multicolumn{1}{c}{32} & \multicolumn{1}{c}{64} & \multicolumn{1}{c}{128} & \multicolumn{1}{c}{256} & \multicolumn{1}{c}{32} & \multicolumn{1}{c}{64} & \multicolumn{1}{c}{128} & \multicolumn{1}{c}{256} & \multicolumn{1}{c}{32} & \multicolumn{1}{c}{64} & \multicolumn{1}{c}{128} & \multicolumn{1}{c}{256}\\
\midrule
\multicolumn{17}{l}{\textit{Existing Methods (likelihood-dependent)}} \\
LOSS & 0.64 & 0.62 & 0.67 & 0.69 & 0.61 & 0.57 & 0.62 & 0.64 & 0.64 & 0.61 & 0.65 & 0.68 & \textbf{0.55} & 0.50 & 0.56 & 0.59 \\
PPL/zlib & 0.65 & 0.63 & 0.68 & 0.69 & 0.61 & 0.58 & 0.64 & 0.65 & 0.64 & 0.62 & 0.67 & 0.70 & \textbf{0.55} & 0.51 & 0.57 & 0.59\\
Lowercase & 0.59 & 0.57 & 0.58 & 0.60 & 0.58 & 0.57 & 0.57 & 0.59 & 0.59 & 0.55 & 0.57 & 0.55 & 0.49 & 0.50 & 0.49 & 0.59 \\
Min-K\% Prob & \textbf{0.67} & \textbf{0.66} & \textbf{0.70} & 0.71 & \textbf{0.62} & 0.60 & 0.67 & 0.67 & \textbf{0.66} &\textbf{0.64} & \textbf{0.69} & 0.71 & 0.51 & 0.50 & 0.56 & 0.58 \\
\midrule
\multicolumn{17}{l}{\textit{Proposed Methods (likelihood-independent)}} \\
SaMIA & 0.55 & 0.60 & 0.63 & 0.77 & 0.55 & 0.63 & 0.66 & \textbf{0.80} & 0.56 & 0.61 & 0.66 & 0.73 & 0.53 & 0.55 & 0.60 & \textbf{0.71} \\
SaMIA*zlib & 0.57 & 0.62 & 0.66 & \textbf{0.79} & \textbf{0.62} & \textbf{0.68} & \textbf{0.70} & \textbf{0.80} & 0.59 & 0.63 & 0.68 & \textbf{0.74} & \textbf{0.55} & \textbf{0.59} & \textbf{0.63} & 0.69 \\
\Xhline{3\arrayrulewidth}
\end{tabular}
\caption{AUC score of the proposed and existing methods when performing MIA on various LLMs using WikiMIA. The best AUC score of each column is \textbf{bolded}.}
\label{tab:auc}
\end{table*}

\paragraph{Models} We evaluate the performance of SaMIA as well as existing MIA methods against LLMs including OPT-6.7B~\cite{zhang2022opt}, Pythia-6.9B~\cite{biderman2023pythia}, LLaMA-2-7B~\cite{touvron2023llama} and GPT-J-6B~\cite{gpt-j}.
Their checkpoints are publicly available on Huggingface\footnote{\url{ https://huggingface.co/}}.
We present an exposition regarding the training data for the LLMs.
\begin{itemize}
  \item \textbf{OPT}~\cite{zhang2022opt}: OPT uses the dataset containing the Pile,\footnote{\url{https://huggingface.co/datasets/EleutherAI/pile}}, which comprises 800GB of text data. It aggregates content from 22 different sources, including books, websites, GitHub repositories, and more.
  Moreover, PushShift.io Reddit~\cite{Baumgartner2020ThePR} is used for pre-training.
  \item \textbf{Pythia}~\cite{biderman2023pythia}: Pythia is trained exclusively on the Pile, similar to OPT.
  \item \textbf{LLaMA-2}~\cite{touvron2023llama}: LLaMA-2 employs English CommonCrawl, C4, Github, Wikipedia, Books, ArXiv, and StackExchange as pre-training datasets.
  \item \textbf{GPT-J}~\cite{gpt-j}: GPT-J uses the Pile for pre-training.
  All corpora were previously collected or filtered to contain predominantly English text, but a small amount of non-English data is still present within the corpus via CommonCrawl.
\end{itemize}

\paragraph{Implementation Details} For both SaMIA and SaMIA*zlib, we utilize ROUGE-1 and set the number of candidate texts to 10.
Analyses of the influence of these factors are detailed in~\autoref{sec:analysis}.
During generation, the hyper-parameters of all models are fixed as \verb|temperature=1.0|, \verb|max_length=1024|, \verb|top_k=50|, \verb|top_p=1.0|.
For existing approaches, we report the scores obtained from our experiments.

\paragraph{Benchmark} Following existing work~\cite{shi2023detecting}, we adopt WikiMIA as the benchmark.
The benchmark contains texts from Wikipedia event pages.
Texts from pages created after 2023 are considered unleaked data (i.e., text excluded from the training data of LLMs), and those from pages created before 2017 are considered leaked data (i.e., text included in the training data of LLMs).

\paragraph{Evaluation Metrics} 
WikiMIA groups texts by length: 32, 64, 128, and 256. 
On each group, we evaluate the performance of detection methods via the True Positive Rate (TPR) and the False Positive Rate (FPR).
We plot the ROC curve to measure the trade-off between TPR and FPR at each threshold $\tau$ and report the AUC score, i.e., the area under the ROC curve, and TPR@10\%FPR, i.e., TPR when FPR is 10\%~\cite{mattern-etal-2023-membership, shi2023detecting}. 

\paragraph{Computational Environment} All experiments are carried out on a single Tesla V100 16GB GPU. We first generate candidate texts using target LLMs and then conduct MIA with existing and proposed methods.
For each LLM, the inference process for the entire WikiMIA dataset takes approximately 48 GPU hours.
Due to the computation cost of running inferences for multiple LLMs, we report the results based on a single run for each model.

\subsection{Main Results}

\begin{figure*}[t]
    \centering
    \begin{subfigure}[t]{0.47\textwidth}
    \centering
    \includegraphics[width=1.0\textwidth]{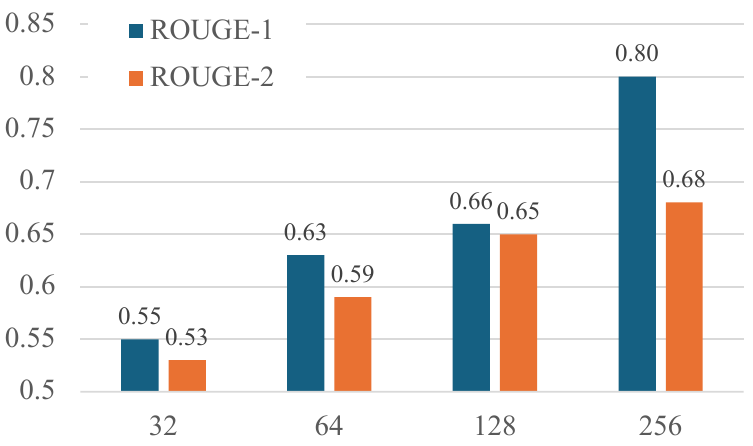}
    \caption{OPT-6.7B.}
    \end{subfigure}
    \begin{subfigure}[t]{0.47\textwidth}
    \centering
    \includegraphics[width=1.0\textwidth]{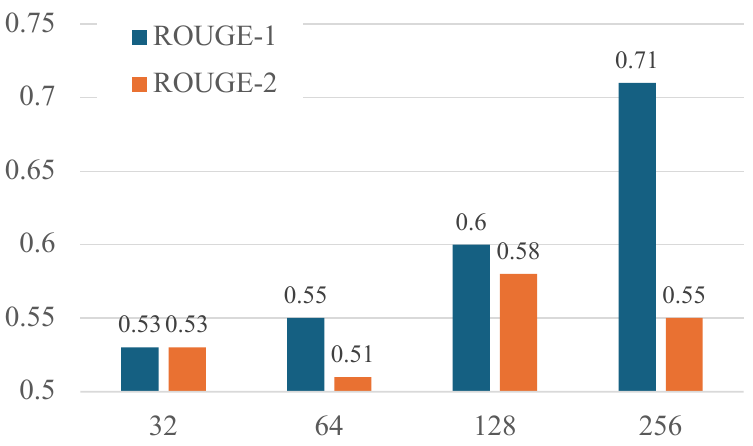}
    \caption{LLaMA-2-7B.}
    \end{subfigure}
    \caption{AUC scores of SaMIA on WikiMIA when using ROUGE-1 and ROUGE-2.}
    \label{fig:ngram}
\end{figure*}

The macro average of AUC scores and TPR@10\%FPR among all groups are reported in \autoref{tab:macro}.
AUC scores of the existing and proposed detection methods for each text length group are reported in~\autoref{tab:auc}.

\paragraph{On average, SaMIA*zlib exhibits state-of-the-art level performance without accessing the likelihood.}
As shown in~\autoref{tab:macro}, when comparing the macro average score of each method, we observe that SaMIA*zlib reaches state-of-the-art performance on OPT-6.7B and LLaMA-2-7B.
For another two LLMs, namely GPT-J-6B and Pythia-6.9B, the performance of SaMIA*zlib is also close to the state-of-the-art method, i.e., Min-K\% Prob.
The high performance is achieved without access to likelihoods computed from LLMs, highlighting the superiority of our proposed method.
While SaMIA*zlib can be applied to any LLMs, existing methods are limited to LLMs with accessible likelihoods.
Thus, the superiority of our proposed method includes not only better performance but also wider usage.

\paragraph{SaMIA outperforms existing methods when the target text is long.}
When evaluating on the group with a text length of 256, we observe SaMIA outperforming all existing methods.
For groups with shorter text lengths, SaMIA still performs comparably or better than its competitors.
Notably, while all existing methods base their detection on the likelihood yielded from LLMs, SaMIA does not rely on the likelihood.
SaMIA, therefore, has its superiority over other methods in achieving competitive or even better performance using less information. 

\paragraph{SaMIA*zlib further improves SaMIA regardless of the text length.}
Incorporating SaMIA with zlib compression entropy further boosts the detection performance for all models, regardless of the length of target texts.
The observation indicates that zlib compression benefits leakage detection by compressing the influence of repeated substrings.

\section{Analysis}
\label{sec:analysis}
This section investigates how each factor will affect the performance of SaMIA(*zlib).
Experiments are conducted on SaMIA without zlib compression, and we believe the findings also apply to SaMIA*zlib.
The analyses are based on OPT-6.7B and LLaMA-2-7B.

\begin{table}[t]
    \centering
    \small
    \begin{tabular}{lccccc}
    \Xhline{3\arrayrulewidth}
    Length & 32 & 64 & 128 & 256 & \textbf{Avg.} \\
    \midrule
    \multicolumn{6}{l}{\textit{OPT-6.7B}} \\
    SaMIA\textsubscript{\textit{Rec.}}  & \textbf{0.55} & \textbf{0.63} & \textbf{0.66} & \textbf{0.80} & \textbf{0.66} \\
    SaMIA\textsubscript{{\textit{Prec.}}} & 0.52 & 0.51 & 0.54 & 0.45 & 0.51 \\
    \midrule
    \multicolumn{6}{l}{\textit{LLaMA-2-7B}} \\
    SaMIA\textsubscript{{\textit{Rec.}}}  & \textbf{0.53} & \textbf{0.55} & \textbf{0.60} & \textbf{0.71} &\textbf{0.60} \\
    SaMIA\textsubscript{{\textit{Pre.}}} & 0.52 & 0.48 & 0.56 & 0.58 & 0.54 \\ 
    \Xhline{3\arrayrulewidth}
    \end{tabular}
    \caption{AUC scores of SaMIA on WikiMIA when surface-level similarity is evaluated by $n$-gram recall and precision. SaMIA\textsubscript{\textit{Rec.}} represents the default SaMIA based on $n$-gram recall and SaMIA\textsubscript{\textit{Prec.}} represents the variant of SaMIA based on $n$-gram precision.}
    \label{tab:precision}
\end{table}

\subsection{ROUGE-1 v.s. ROUGE-2}

\begin{table*}
\centering
\footnotesize
\begin{tabular}{lcccccccccc}
\Xhline{3\arrayrulewidth}
Samples & 1 & 2 & 3 & 4 & 5 & 6 & 7 & 8 & 9 & 10 \\
\midrule
OPT-6.7B & 0.61 & 0.64 & 0.65 & 0.65 & 0.65 & 0.66 & 0.66 & 0.66 & 0.66 & 0.66 \\
LLaMA-2-7B & 0.59 & 0.59 & 0.59 & 0.58 & 0.60 & 0.60 & 0.60 & 0.60 & 0.60 & 0.60 \\
\Xhline{3\arrayrulewidth}
\end{tabular}
\caption{AUC scores of SaMIA with different sampling sizes on WikiMIA. Values reported are macro-averages among all target length groups.}
\label{tab:sampling}
\end{table*}

We have reported the performance of SaMIA based on ROUGE-1, i.e., unigram overlaps between generated texts and the reference text, in~\autoref{tab:auc}.
Here, we investigate how varying the size of units used to measure surface-level similarity affects our leakage detection.
Specifically, we evaluate the performance of SaMIA based on ROUGE-2 by setting $n$ to 2 in~\autoref{eq:SaMIA}.

ROUGE-2 differs from ROUGE-1 in modeling bi-gram overlaps~\cite{lin-2004-rouge}.
Intuitively, it is easier for an LLM to output the same word in the reference text than two continuous ones.
Therefore, we presume that a bi-gram match between the generated texts and the reference text is less likely to happen than a unigram match, making ROUGE-2 a less sensitive metric for detection.

The AUC scores of SaMIA based on ROUGE-1 and ROUGE-2 are shown in~\autoref{fig:ngram}.
On most text length groups, the AUC score of SaMIA based on ROUGE-1 surpasses that of ROUGE-2.
For both LLMs, the performance gap between ROUGE-1 and ROUGE-2 becomes evident for the text group with a length of 256.
We thus confirmed that measuring surface-level similarity with unigram overlap is a better choice than bi-gram for SaMIA.

\subsection{Recall-Based Similarity v.s. Precision-Based Similarity}

To measure the surface-level similarity between a generated candidate text $x_\text{cand}$ and the reference text $x_\text{ref}$, SaMIA utilizes a recall-based metric, computing the ratio of $n$-grams correctly recalled from $x_\text{cand}$.
Here, we conduct experiments to test the effectiveness of the recall-based metric against a precision-based one.
For each candidate text $x_\text{cand}$, the precision-based metric evaluates how many $n$-grams within $x_\text{cand}$ overlaps with $x_\text{ref}$ among all generated $n$-grams.
Specifically, we replace the denominator in~\autoref{eq:rougen} with the total number of $n$-grams in $x_\text{cand}$:

{\small
\begin{equation}
\text{Precison-N}(x_\text{cand}, x_\text{ref}) = \frac{\sum_{\text{gram}_n\in x_\text{ref}} \text{Count}_\text{match}(\text{gram}_n)}{\sum_{\text{gram}_n\in x_\text{cand}}\text{Count}(\text{gram}_n)} \label{eq:precisionn}. 
\end{equation}
}
The leakage detection is thus performed by thresholding the averaged Precision-N score among all candidate texts and the reference text.

\autoref{tab:precision} demonstrates the experiment results.
We observe that for both LLMs, leakage detection based on precision yields nearly random performance (AUC$\approx$0.50).
The observation suggests that it is more beneficial to focus on the quantity of information recalled from the reference text than how precisely the reference text is reproduced.
One possible explanation for this can be the difficulty of reproducing the text exactly without introducing any redundant information.

\subsection{Number of Samples}

\begin{figure}[t]
    \centering
    \includegraphics[width=0.47\textwidth]{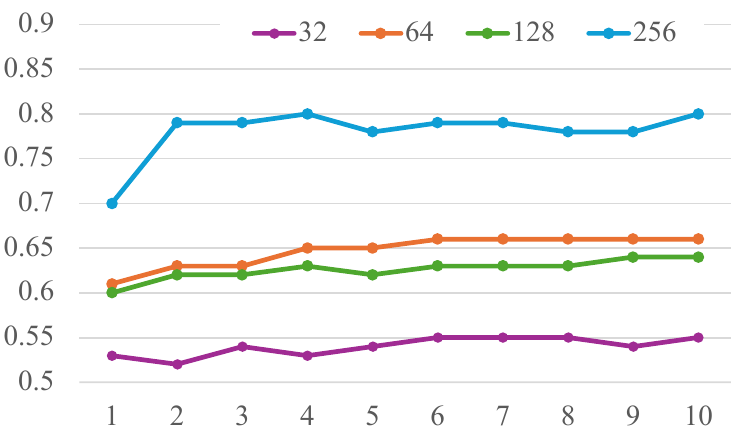}
    \caption{AUC scores of SaMIA on WikiMIA with different sampling sizes. The target LLM for leakage detection is OPT-6.7B.}
    \label{fig:sampling}
\end{figure}

Here, we investigate the effect of sampling size, i.e., the number of generated candidate texts, on SaMIA.
Intuitively, generating more samples provides more information about the target LLM, thus helping better simulate the ``real'' distribution.
The pseudo-likelihood computed from a larger number of samples should, therefore, be closer to the true likelihood obtained from LLMs.
We thus hypothesize that the performance of SaMIA is positively related to the number of samples.

\autoref{tab:sampling} shows how the performance of SaMIA varies with the sampling size.
To analyze the trend more effectively, we report the AUC score for each text length group alongside the changes in sampling size in~\autoref{fig:sampling}.
On OPT-6.7B, while the performance of SaMIA based on a single sample is limited, adding one or two more samples improves the performance for all text-length groups.
However, the improvement fades out when the sampling size exceeds 5.
The observation suggests that our hypothesis is partially correct, but maintaining many samples is unnecessary.
A similar phenomenon can be observed on LLaMA-2-7B, where the performance of SaMIA stabilizes with multiple samples.
Therefore, we conclude that a hyper-parameter search should be performed to discover a balanced sampling size that optimizes both detection accuracy (stability) and inference cost.

\begin{figure}[t]
    \centering
    \includegraphics[width=0.47\textwidth]{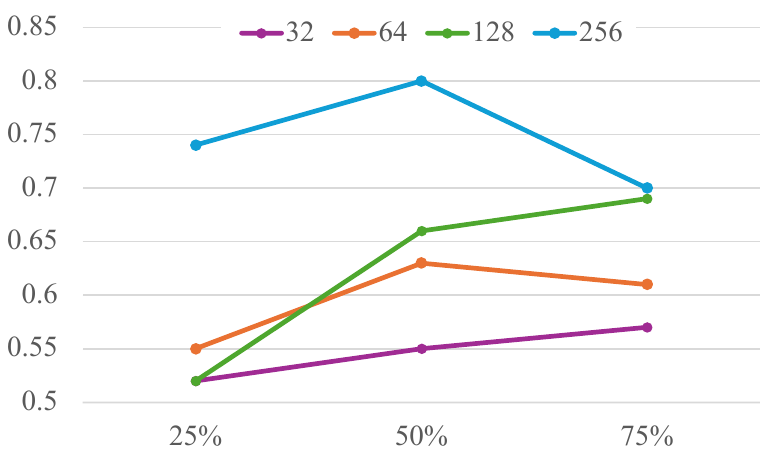}
    \caption{AUC scores of SaMIA on WikiMIA with different prefix ratios. The target LLM for leakage detection is OPT-6.7B.}
    \label{fig:prefix}
\end{figure}

\begin{figure*}
    \centering
    
    \begin{subfigure}[t]{0.45\textwidth}
    \centering
    \includegraphics[width=1.0\textwidth]{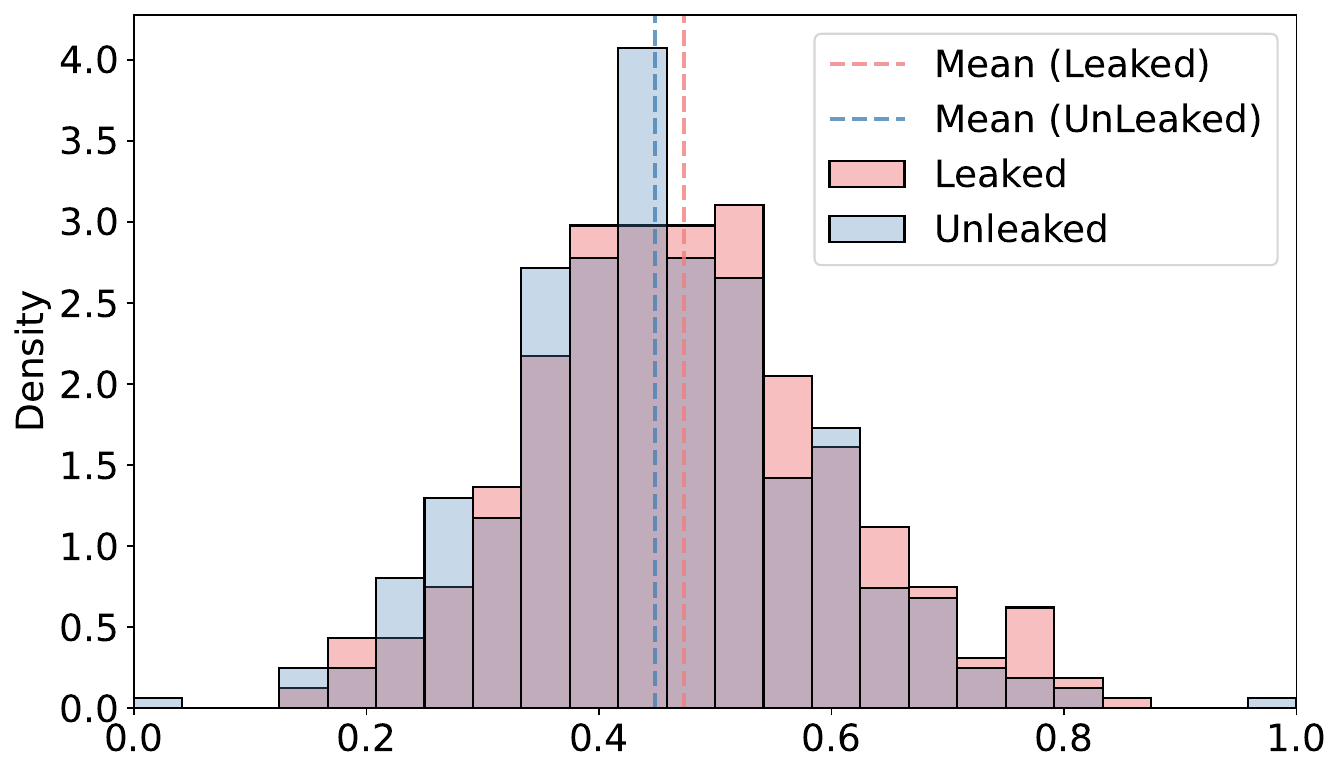}
    \caption{target text length = 32.}
    \end{subfigure}
    \begin{subfigure}[t]{0.45\textwidth}
    \centering
    \includegraphics[width=1.0\textwidth]{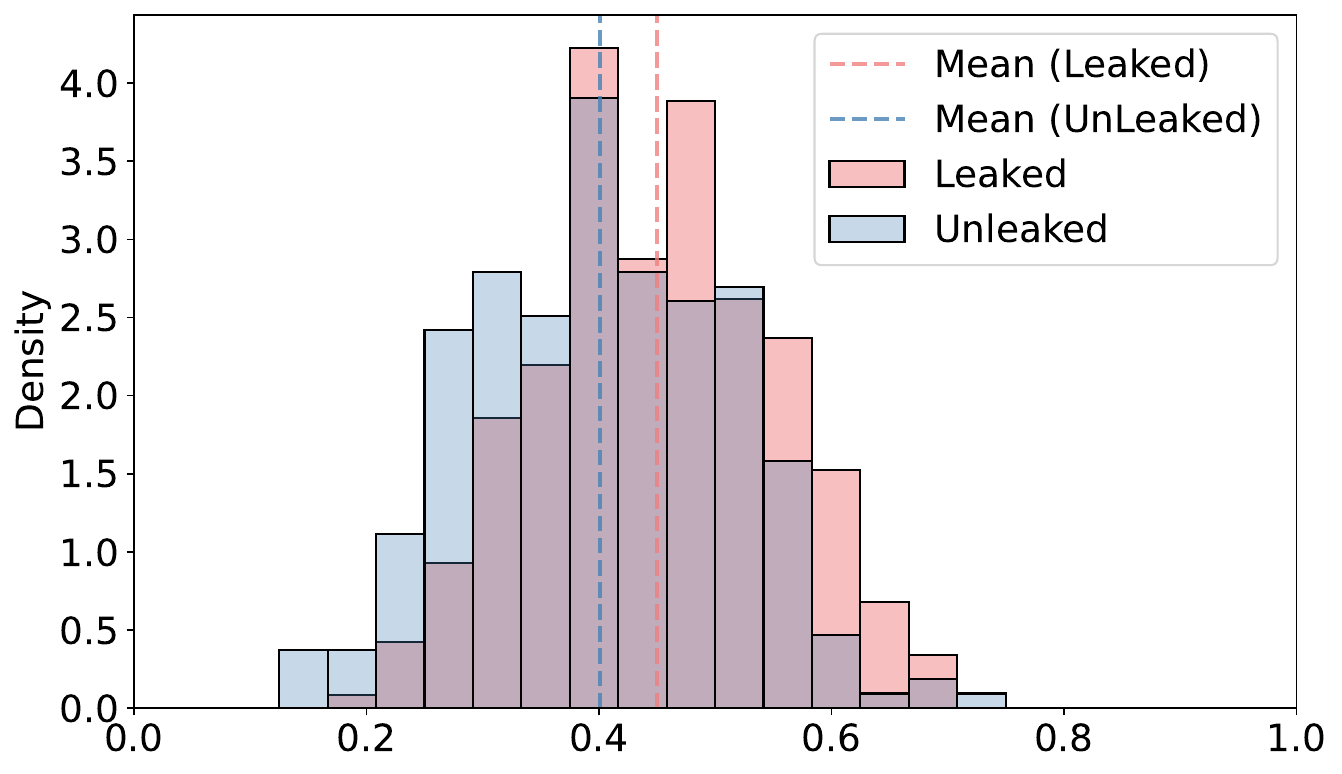}
    \caption{target text length = 64.}
    \end{subfigure}
    
    \begin{subfigure}[t]{0.45\textwidth}
    \centering
    \includegraphics[width=1.0\textwidth]{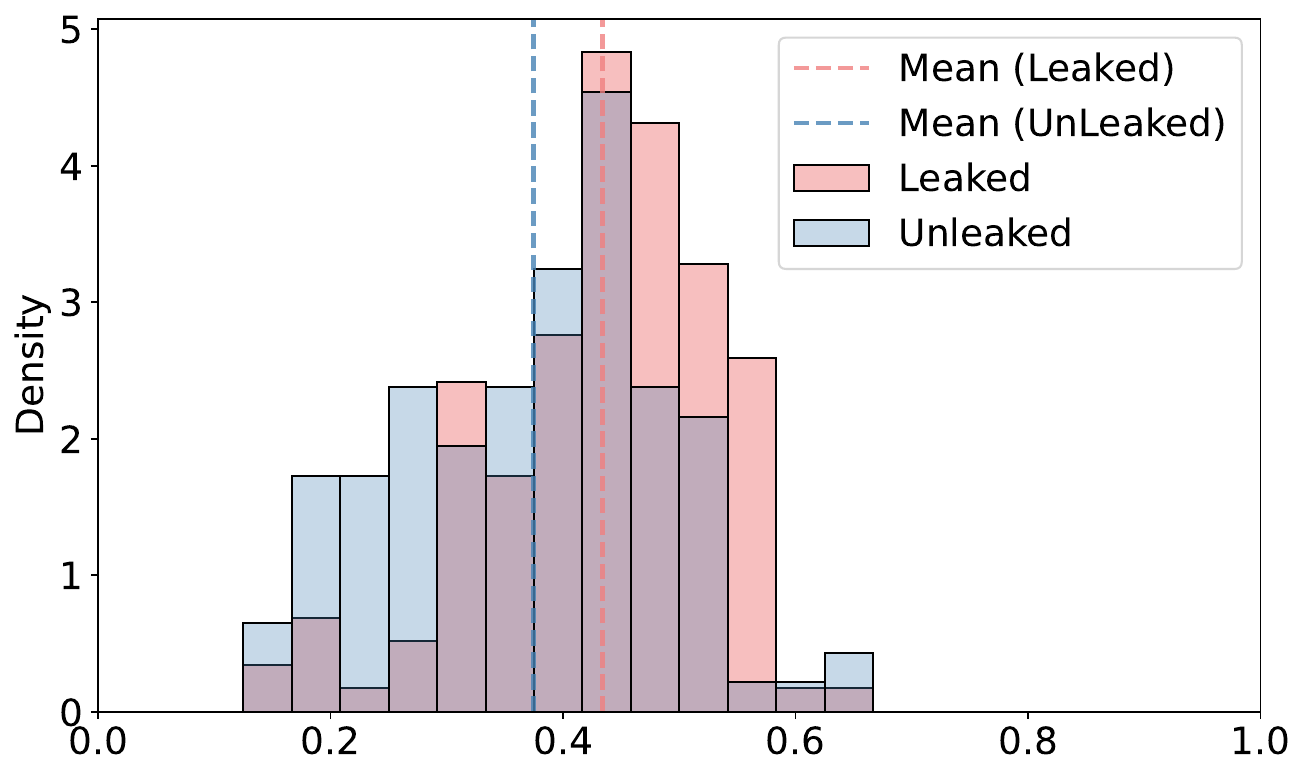}
    \caption{target text length = 128.}
    \end{subfigure}
    \begin{subfigure}[t]{0.45\textwidth}
    \centering
    \includegraphics[width=1.0\textwidth]{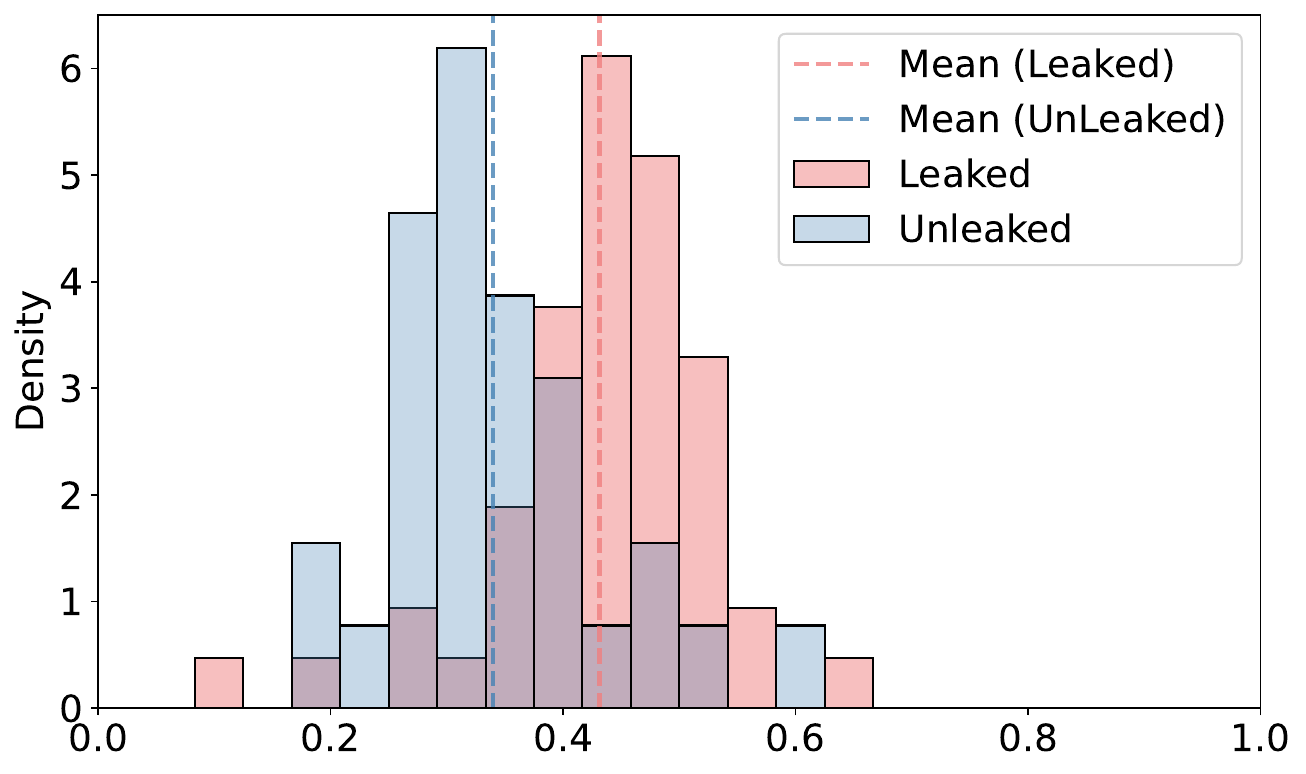}
    \caption{target text length = 256.}
    \end{subfigure}
    
    \caption{ROUGE-1 scores of texts generated from OPT-6.7B, using the original texts in WikiMIA as references. Red bars show the distribution of leaked texts and blue bars show that of unleaked ones.}
    \label{fig:rouge}
\end{figure*}

\subsection{Length of Prefix}

As introduced in~\autoref{sec:spl}, for each target text $x$, we divide it into two halves, where the first half serves as the prefix (i.e., prompt) to generate the second half.
Here, we investigate how SaMIA behaves if provided with a shorter or longer prefix.
A shorter prefix limits the information accessible for the LLM, making it more challenging to recall the preceding context accurately. Consequently, this increases the difficulty of leakage detection.

Specifically, we conduct experiments using 25\%, 50\%, and 75\% of the target text as the prefix, with results shown in~\autoref{fig:prefix}.
In general, compared with the setting where half of the whole text is provided as the prefix, limiting the prefix ratio to 25\% causes performance drops.
However, increasing the prefix ratio from 50\% to 75\% does not consistently improve performance; instead, we observe a notable performance decline in the text group with 256 words.
The result shows that an optimal prefix length falls within the middle range, consistent with the trend described in \citet{yang2024dnagpt}, where the prefix-generated texts from LLMs are used to distinguish between text written by the LLMs and humans.
The observation indicates that, while prefixes provide important hints for leakage detection, a longer prefix does not guarantee better performance.
For text included in the training data, there may exist a soft threshold at which LLMs can effectively recall the context, rendering additional information unnecessary.
We left the exploration of such a soft threshold to future work.

\subsection{Length of Target Text}
\autoref{tab:auc} has demonstrated that as the target text length increases, the performance of SaMIA also improves.
Here, we conduct experiments to help understand the reason.
Given that SaMIA detects leakage based on ROUGE-N (N=1), we investigate how ROUGE-1 scores of leaked and unleaked texts differ for each text length group.

The distribution of ROUGE-1 scores of leaked and unleaked texts for different lengths is shown in~\autoref{fig:rouge}.
With the text length ranging from 32 to 256, the distance between the distributions of leaked and unleaked texts increases.
As the distributions of leaked and unleaked texts move far from each other, they become more separable, resulting in more precise leakage detection via thresholding.
As described in~\autoref{sec:spl}, the first half of each target text serves as the prompt for generating the second half.
Therefore, one possible explanation can be that longer prompts reduce ambiguity, aiding LLMs in better recalling the memory and generating subsequent contents.

\section{Conclusion}

We propose SaMIA, the first likelihood-independent approach for leakage detection based on sampling.
For each target text, a pseudo-likelihood is calculated based on ROUGE-N, where leaked texts yield higher pseudo-likelihood than unleaked ones.
In experiments, SaMIA demonstrates performance on par with existing likelihood-dependent methods, even outperforming those when the target text length is long.
While existing methods can only be applied to leakage detection on LLMs with accessible likelihoods, SaMIA can be applied to any LLMs.
In the future, we plan to explore ways to improve the accuracy of leakage detection, especially for short texts.

\section*{Limitations}

Typically, LLMs with non-public training data like ChatGPT and Gemini would be the target of the MIA.
However, since it is not possible to prepare both trained and untrained data, in this study we targeted LLMs with known training data.
In our experiment, we only conducted the experiment on a single dataset~\cite{shi2023detecting}, and investigating the robustness of SaMIA across different tasks is a future challenge.

\section*{Ethical Considerations}

LLMs are known to have issues regarding fairness, toxicity, and social bias~\cite{oba2023contextual,anantaprayoon2023evaluating,kaneko2024gaps,kaneko2024eagle,kaneko2024evaluating}.
In this paper, the experiments were conducted using existing data and existing models, so there are no new ethical concerns from a resource perspective.
On the other hand, when sampling the output of the LLM to compute the SPL, there is a possibility of generating text that is related to those concerns. 
However, since our method ultimately outputs a score rather than text, this does not pose a problem.

\section*{Acknowledgements}
This work was supported by JSPS KAKENHI Grant Number 19H01118.

\bibliography{custom}

\newpage
\appendix
\section{WikiMIA: Statistics}

Experiments in this work are based on WikiMIA~\cite{shi2023detecting}.
The dataset is a benchmark for evaluating membership inference attack methods and comprises Wikipedia event pages from 2017 to 2023.
The texts are divided into 4 groups with different lengths, namely 32, 64, 128, and 256.
We regard texts with a timestamp earlier than 2023 as data included in the training data for LLMs (i.e., leaked data) and those with a timestamp later than 2023 as data excluded from the training data (i.e., unleaked data).  
\autoref{tab:wikimia} details the statistics of WikiMIA.

\begin{table}[t]
    \centering
    \begin{tabular}{lccc}
    \Xhline{3\arrayrulewidth}
         Length &  Leaked & Unleaked & Total \\
         \midrule
         32 & 387 & 389 & 776\\
         64 & 284 & 258 & 542\\
         128 & 139 & 11 & 250 \\
         256 & 51 & 31 & 82\\
         Total & 861 & 789 & --\\
    \Xhline{3\arrayrulewidth}
    \end{tabular}
    \caption{Number of leaked/unleaked instances in each text length group of WikiMIA.}
    \label{tab:wikimia}
\end{table}

\end{document}